\def\year{2020}\relax
\documentclass[letterpaper]{article} % DO NOT CHANGE THIS
\usepackage{aaai20}  % DO NOT CHANGE THIS
\usepackage{times}  % DO NOT CHANGE THIS
\usepackage{helvet} % DO NOT CHANGE THIS
\usepackage{courier}  % DO NOT CHANGE THIS
\usepackage[hyphens]{url}  % DO NOT CHANGE THIS
\usepackage{graphicx} % DO NOT CHANGE THIS
\usepackage{graphicx}  % DO NOT CHANGE THIS
\usepackage{algorithm}
\usepackage{algpseudocode}
\usepackage{subfigure,underoverlap}
\usepackage{amsfonts}       % blackboard math symbols
\usepackage{multirow}

\nocopyright
\title{A Kernel to Exploit Informative Missingness in Multivariate Time Series from EHRs}
\author{Karl {\O}yvind Mikalsen\thanks{Corresponding author. https://machine-learning.uit.no/}\textsuperscript{\rm ,1,2} , Cristina Soguero-Ruiz \textsuperscript{\rm 1,3}, Robert Jenssen \textsuperscript{\rm 1} 
\\ % All authors must be in the same font size and format. Use \Large and \textbf to achieve this result when breaking a line
 \textsuperscript{\rm 1} Dept. of Physics and Technology, UiT The Arctic University of Norway, NO-9037 Troms{\o}, Norway\\
   \textsuperscript{\rm 2} University Hospital of North-Norway, \textsuperscript{\rm 3} Rey Juan Carlos University, Fuenlabrada, Spain\\
karl.o.mikalsen@uit.no, cristina.soguero@urjc.es, robert.jenssen@uit.no  % email address must be in roman text type, not monospace or sans serif
}
 
\begin{document}

\maketitle

\begin{abstract}
A large fraction of the electronic health records (EHRs) consists of clinical measurements collected over time, such as lab tests and vital signs, which provide important information about a patient's health status. These sequences of clinical measurements are naturally represented as time series, characterized by multiple variables and large amounts of missing data, which complicate the analysis. In this work, we propose a novel kernel which is capable of exploiting  both the information from the observed values as well the information hidden in the missing patterns in multivariate time series (MTS) originating e.g. from EHRs.
The kernel, called TCK$_{IM}$, is designed using an ensemble learning strategy in which the base models are novel mixed mode Bayesian mixture models which can effectively exploit informative missingness without having to resort to imputation methods. Moreover, the ensemble approach ensures robustness to hyperparameters and therefore TCK$_{IM}$ is particularly well suited if there is a lack of labels - a known challenge in medical applications.
Experiments on three real-world clinical datasets demonstrate the effectiveness of the proposed kernel.
\end{abstract}

\section{Introduction}
The widespread growth of electronic health records (EHRs) has generated vast amounts of clinical data.  Normally, an encounter-based patient EHR is longitudinal and contains clinical notes, diagnosis codes,  medications, laboratory tests and vital signs, etc., which can be represented as multivariate time series (MTS). As a consequence, EHRs contain valuable information about the clinical observations depicting  both patients’ health and care provided by physicians. 
However, the longitudinal and heterogeneous data sources make EHR analysis difficult from a computational perspective. In addition, the EHRs are often subject to  a lack of completeness, implying that the MTS extracted from EHRs often contain massive \textit{missing data}~\cite{sharafoddini2019new}. 
Missing data might, however, occur for different reasons. It could be that the physician orders lab tests, but because of an error some results are not recorded. On the other hand, it can also happen that the physician decides to not order lab tests because he thinks the patient is in good shape. In the first case, the missingness is ignorable, whereas in the latter case, the missing values and patterns potentially can contain rich information about the patient's diseases and clinical outcomes. Efficient data-driven systems aiming to extract knowledge, perform predictive modeling, etc., must be capable of capturing this information.

Traditionally, missingness mechanisms have been divided into missing completely at random (MCAR), missing at random  (MAR) and missing not at random (MNAR). The main difference between these mechanisms consists in whether the missingness is ignorable (MCAR and MAR) or non-ignorable (MNAR)~\cite{donders2006review,rubin1976inference}. 
This traditional description of missingness mechanisms is, however, not always sufficient in medical applications as the missing patterns might be correlated with additional variables, such as e.g. a disease. This means that the distribution of the missing patterns for patients with a particular disease might be different than the corresponding distribution for patients without the disease, i.e. the missingness is informative~\cite{ghorbani2018embedding,wells2013strategies}. 

Several methods have been proposed to handle missing data in MTS~\cite{little2014statistical,schafer2002missing}. A simple approach is to create a complete dataset by discarding the MTS with missing data. Alternatively, one can do simple imputation of the missing values, e.g. using the last observation carried forward scheme (impute the last non-missing value for the following missing values)~\cite{shao2003last}, zero-value imputation (replace missing values with zeros) or mean-value imputation (missing values are replaced with the mean of the observed data)~\cite{zhang2016missing}. % Other so-called single imputation methods exploit machine learning based  methods such as k-nearest neighbors, random forest, multilayer perceptrons, self-organizing maps, recurrent neural  networks and regression-based imputation~\cite{jerez2010missing}.
A common limitation of these approaches is that they lead to additional bias, loss of precision, and they ignore uncertainty associated with the missing values~\cite{donders2006review}. 
This problem is to some extent solved via multiple imputation methods, i.e. by creating multiple complete datasets using single imputation independently each time. Then, by training a classifier using an ensemble learning strategy, one can improve the performance compared to simple imputation. However, this imputation procedure is an ad-hoc solution as it is performed independently of the rest of the analysis and it ignores the potential predictive value of the missing patterns~\cite{MA2018297}.

Due to the limitations of imputation methods, several research efforts have been devoted over the last years to deal with missing data in MTS using alternative strategies~\cite{8489716,bianchi2019learning,DBLP:journals/corr/ChePCSL16,Ghassemi,li2019vs,pmlr-v56-Lipton16,mikalsen2016learning,shukla2018interpolation}. Prominent examples are kernels, i.e. positive semi-definite time series similarity measures,  such as the \textit{learned pattern similarity} (LPS)~\cite{baydogan2016time} and the \textit{time series cluster kernel} (TCK)~\cite{mikalsen2018time} that can naturally deal with missing data.
The former generalizes autoregressive models to local autopatterns, which capture the local dependency structure in the time series, and uses an  ensemble learning (random forest) strategy in which a bag-of-words representation is created from the output of the leaf-nodes for each tree. %The kernel is then computed using a histogram intersection kernel~\cite{barla2003histogram} from the concatenated representation of the outputs of all leaf nodes.
TCK is also based on an ensemble learning approach  and shares many properties with LPS.  It is designed using an ensemble learning approach in which Bayesian mixture models form the base models. 
However, while LPS exploits the inherent missing data handling abilities of decision trees, TCK is a likelihood-based approach in which the incomplete  dataset is analysed using  maximum a posteriori  expectation-maximization. %Both of these approaches assume that the missingness is ignorable.
An advantage of these methods, compared to e.g. multiple imputation that requires a careful selection of imputation model and parameters~\cite{schafer2002missing}, is that the missing data are handled automatically and no additional tasks are left to the designer. Additionally, since the methods are based on ensemble learning, they are robust to hyperparameter choices. In particular, these properties are important in unsupervised settings, which frequently occur in medical applications where manual label annotation of large datasets often is not feasible~\cite{Halpernocw011,MIKALSEN2017105}. 

A shortcoming of these kernel methods is, however, that they cannot exploit informative missing patterns, which frequently occur in medical MTS, and unbiased predictions are only guaranteed for ignorable missingness as MAR is an underlying assumption.
Recently, several studies have focused on modeling the informative or nonignorable missigness by analyzing the observed values as well as the indicators of missingness, concluding that the missing patterns can add more insights beyond the observed values~\cite{agniel2018biases,DBLP:journals/corr/ChePCSL16,pmlr-v56-Lipton16,sharafoddini2019new}. 
In this work, we present a novel time series cluster kernel, TCK$_{IM}$, that also represents the missing patterns using binary indicator time series. By doing so, we obtain MTS consisting of both continuous and discrete attributes. However,  we do not only concatenate the binary MTS to the real-valued MTS and analyse these data in a naive way. Instead, we take a statistically principled Bayesian approach~\cite{little2014statistical,MA2018297} and model the missingness mechanism more rigorously by introducing novel mixed mode Bayesian mixture models, which can effectively exploit information provided by the missing patterns as well as the temporal dependencies in the observed MTS.  The mixed mode Bayesian mixture models are then used as base models in an ensemble learning strategy to form the TCK$_{IM}$ kernel. Experiments on three real-world datasets of patients described by longitudinal EHR data, demonstrate the effectiveness of the proposed method.
%\textbf{TO do: Try to place the bayesian modeling in the context of other lines of work. Have people done similar stuff? What is different in the core modeling idea?}

%The remainder of this paper is organized as follows. Sec.~\ref{Sec:TCK_IM} presents the proposed kernel. Experiments on three EHR datasets are described in Sec.~\ref{sec: experiments}, whereas results and discussions are presented in Sec.~\ref{Sec:Results}.% Conclusions are presented in Sec.~\ref{Sec:Conclusion}.

\section{Time series cluster kernel to exploit informative missingness}
\label{Sec:TCK_IM}
Here we present the proposed TCK$_{IM}$ kernel.   The kernel is learned using an ensemble learning strategy, i.e. by training individual base models which are combined into a composite kernel in the end. 
As base model we introduce a novel mixed mode Bayesian mixture model. Before we provide the details of this method, we describe the notation used throughout the paper.

\textbf{Notation} We define a  multivariate time series (MTS) $X$  as a finite combination of univariate time series (UTS) $x_v$ of length $T$, i.e.
$
X = \{ x_v \in \mathbb{R}^T \: | \: v = 1,2,\dots,V\}.
$
The dimensionality of the MTS $X$ is the same as the number of UTS, $V$, and the length of $X$ is the same as the length $T$ of the UTS $x_v$. A $V$--dimensional MTS, $X$, of length $T$ can be represented as a matrix in $\mathbb{R}^{V \times T}$. Given a dataset of $N$ MTS, we denote $X^{(n)}$ as the $n$-th MTS. 
In a dataset of $N$ incompletely observed MTS, the $n$-th MTS is denoted by the pair $U^{(n)} = (X^{(n)}, R^{(n)})$, where $R^{(n)}$ is a binary MTS 
with entry $r_v^{(n)}(t) = 0$ if the realization $x_v^{(n)}(t)$ is missing and $r_v^{(n)}(t) = 1$ if it is observed.

\paragraph{Mixed mode Bayesian mixture model}
Let $ U = (X, R)$ be a MTS generated from two modes, where $X$ is a V-variate real-valued MTS ($ X \in \mathbb{R}^{V \times T}$) and $R$ is a V-variate binary MTS ($ R \in \{0,1\}^{V \times T}$). In the mixture model it is assumed that $U$ is generated from a finite mixture density
\begin{equation}
\textstyle
    p_u(U \: | \: \Phi, \Theta ) = \sum_{g=1}^G \theta_g p_{u_g}( U \: | \: \phi_g),
\end{equation}
where $G$ is the number of components, 
$p_{u_g}$ is the density of the components parametrized by $\Phi = (\phi_1, \dots, \phi_G)$,
and $\Theta = (\theta_1, \dots, \theta_G) $ are the mixing coefficients, $0 \leq \theta_g \leq 1$ and $ \sum_{g=1}^G \theta_g = 1$. 
We formulate the mixture model in terms of a latent random variable $ Z $, described via the one-hot vector $Z = (Z_1,\dots,Z_G)$ with marginal distribution given by 
$
p_z(Z \: | \: \Theta ) = \prod_{g=1}^G  \theta_g^{Z_g}. 
$
The latent variable $Z$ describes which cluster component the MTS $U$ belongs to, i.e. $Z_g = 1$ if $U$ belongs to cluster component $g$ and $Z_g = 0 $ otherwise.
The conditional is given by
$
    p_{u|z}(U \: | Z, \: \Phi)  = \prod_{g=1}^G p_{u_g}( U \: | \: \phi_g)^{Z_g},
$
and therefore it follows that the joint distribution is given by
\begin{align}
%\small
%\textstyle
    p_{u,z}(U, Z \: | \:  \Phi, \Theta) %&=  \sum_{g=1}^G P(W \: | \:  Z_g, \xi_g) P (Z_g \: | \: \xi_g) \nonumber \\
    %&= \Phi_{g=1}^G \left[  P(W \: | \:   \xi_g) P (Z_g \: | \: \xi_g)  \right]^{Z_g}   \\
     =p_{u|z}(U \: | Z, \: \Phi)  p_z(Z \: | \: \Theta ) \nonumber \\
     = \textstyle \prod_{g=1}^G\left[  p_{u_g}(U \: | \:   \phi_g) \theta_g \right]^{Z_g}.
\end{align}
We further assume that the parameters of each component are given by $\phi_g = (\mu_g, \Sigma_g, \beta_g) $ and
%$
\begin{equation} 
    p_{u_g}(U \: | \:   \phi_g) = p_{x|r}(X \: | \: R, \mu_g, \Sigma_g ) p_r( R \: | \: \beta_g), 
\end{equation}
%$
where
$ p_{x|r} $ is a density function given by
\begin{equation}  \label{eq: diag gmm density}
\small
\textstyle
    p_{x|r}(X \: | \: R, \mu_g, \Sigma_g  ) =  \prod_{v=1}^V \prod_{t=1}^T  \mathcal{N} (x_v(t) \: | \: \mu_{gv}(t), \sigma_{gv})^{r_v(t) },
\end{equation}
%say something more than just naive assumption, we are 
where $\mu_g = \{ \mu_{gv} \in  \mathbb{R}^T \: | \: v = 1,...,V\}$ is a time-dependent mean, and
$\Sigma_g = diag\{\sigma_{g1}^2,...,\sigma_{gV}^2\}$ is
a diagonal covariance matrix in which $\sigma_{gv}^2$ is the variance of attribute $v$. Hence, the covariance is assumed to be constant over time.
$p_r $ is a probability mass given by
\begin{equation}  \label{eq: BMM prob mass}
%\small
\textstyle
    p_r( R \: | \: \beta_g) =  \prod_{v=1}^V \prod_{t=1}^T \beta_{gvt}^{r_v(t)} (1 - \beta_{gvt})^{1-r_v(t)}, 
\end{equation}
where $\beta_{gvt} \in [0,1]$. The idea with this formulation is to use the Bernoulli term $p_r $ to capture information from the missing patterns and  $ p_{x|r} $ to capture the information from the observed data.

Using Bayes' theorem we compute the conditional probability of $Z$ given $U$, $ P(Z_g = 1 | U,  \Phi, \Theta ) $,
\begin{align} \label{eq: p(z|x) posterior}
\pi_{g} 
= \frac{\theta_g  p_{x|r}(X  |  R, \mu_g, \Sigma_g)  p_r( R  |  \beta_g)  }{\sum_{g=1}^G \theta_g p_{x|r}(X  |  R, \mu_g, \Sigma_g)  p_r( R |  \beta_g)}.
%= \footnotesize{\frac{ \theta_g \prod\limits_{v=1}^V \prod\limits_{t=1}^T  \left[ \mathcal{N} (x_v(t) \: | \: \mu_{gv}(t), \sigma_{gv}) \beta_{gvt}\right]^{r_v(t)}  (1 - \beta_{gvt})^{1-r_v(t)} }{\sum\limits_{g=1}^G \theta_g \prod\limits_{v=1}^V \prod\limits_{t=1}^T  \left[ \mathcal{N} (x_v(t) \: | \: \mu_{gv}(t), \sigma_{gv}) \beta_{gvt}\right]^{r_v(t)}  (1 - \beta_{gvt})^{1-r_v(t)}}}.
\end{align}
%The parameters of the DiagGMM are learned using a maximum a posteriori expectation maximization algorithm, as described in \cite{mikalsen2017time}. 
To improve the capability of handling missing data, a Bayesian extension is introduced where  informative priors are put over the parameters of the normal distribution as well as the Bernoulli distribution. This enforces that the cluster representatives become smooth over time even in the presence of large amounts of missing data and that the parameters of clusters with few MTS are similar to the overall mean. Towards this end, a kernel-based Gaussian prior is defined for the mean,
\begin{equation} \label{eq: mu}
%\small
%\textstyle
p_{\mu}(\mu_{gv}) = \mathcal{N} \left(\mu_{gv} \: | \: m_{v}, \: S_{v}\right),
\end{equation}
where $m_{v}$ are the empirical means and $
S_{v} = s_{v} \mathcal{K},
$ 
are the prior covariance matrices. 
$s_{v}$ are empirical standard deviations and $\mathcal{K}$ is a kernel matrix, whose elements are
$
\mathcal{K}_{tt'} = b_0 \exp (-a_0(t-t')^2), \quad t, \, t' = 1,\dots,T,
$
with $a_0$, $b_0$ being user-defined hyperparameters.  For the standard deviation $\sigma_{gv}$, an inverse Gamma distribution prior is introduced
\begin{align} \label{eq: sigma}
p_{\sigma}(\sigma_{gv}) \propto \sigma_{gv}^{-N_0} e^{- N_0 s_v/2 \sigma_{gv}^2 },
\end{align}
where $N_0$ is a hyperparameter. Further, we put a Beta distribution prior on $\beta_{gvt}$
\begin{equation} \label{eq: beta}
    p_{\beta}(\beta_{gvt}) \propto \beta_{gvt}^{c_0 - 1} (1- \beta_{gvt})^{d_0 - 1}.
\end{equation}
where $ c_0 $, $d_0$ are hyperparameters. 
We let $\Omega = \{ a_0, b_0, c_0, d_0, N_0\}$ denote the set all of hyperparameters.

Given a dataset $\{U^{(n)}\}_{n=1}^N$, we estimate the parameters $\{ \Phi, \Theta \}$ using maximum a posteriori expectation maximization (MAP-EM)~\cite{dempster1977maximum}. 
The Q-function is computed as follows
\small{
\begin{align} \label{eq: Q}
    Q &= \mathbb{E}_{Z \: | \: U, \: \Theta, \: \Phi } \left[ \log \left( p_{u, z}(U, Z \: | \:  \Phi, \Theta) p(\Phi) \right) \right] \nonumber \\
 %&= \mathbb{E}_{Z \: | \: U, \: \Theta, \: \Phi } \left[ \log \left( p_{u, z}(U, Z \: | \:  \Phi, \Theta) \right) \right] + \log p(\Phi)  \nonumber \\
    &=   \log p(\Phi)  +  \textstyle \sum_{n, g}  \log [p_{u}(U^{(n)}  |  Z^{(n)}_g = 1,  \Phi)  p(Z^{(n)}_g  |  \Theta ) ] \pi_{g}^{(n)} \nonumber \\
    &= \textstyle \sum_{g} \log \left( p_{\mu}(\mu_{g})  p_{\sigma}(\sigma_{g}) p_{\beta} (\beta_{g}) \right) \nonumber \\
&\quad + \textstyle \sum_{n, g}  \log \big[ p_{x | r}(X^{(n)}  |  R^{(n)}, \mu_g, \Sigma_g) p_r( R^{(n)}  | \beta_g) \theta_g \big]   \pi_{g}^{(n)},  %\nonumber \\
\end{align}
}
\normalsize
where $p_{x|r}$ is given by Eq.~\eqref{eq: diag gmm density}, $p_r$ by Eq.~\eqref{eq: BMM prob mass}, $p_{\mu}$ by Eq.~\eqref{eq: mu}, $p_{\sigma}$ by Eq.~\eqref{eq: sigma} and $p_{\beta}$ by Eq.~\eqref{eq: beta}.

The E-step in MAP-EM is the same as in maximum likelihood EM and consists in updating the posterior (Eq.~\eqref{eq: p(z|x) posterior}) using the current parameter estimates, whereas the M-step consists in maximizing the Q-function (Eq.~\eqref{eq: Q}  wrt. the parameters $\{ \Phi, \Theta \}$.
I.e., 
\begin{equation}
\small
%\textstyle
    \{ \Phi^{(m+1)}, \Theta^{(m+1)} \} = \arg \max_{\{ \Phi, \Theta \}} Q(  \Phi, \Theta \: | \:  \Phi^{(m)}, \Theta^{(m)})% \nonumber \\
   % &= \arg \max_{\{ \Phi, \Theta \}} \mathbb{E}_{Z \: | \: U, \: \Theta^{(m)}, \: \Phi^{(m)} } \left[ \log \left( p_{u, z}(U, Z \: | \:  \Phi, \Theta) p(\Phi) \right) \right] 
\end{equation}
Computing  the derivatives of $Q$ with respect to the parameters $ \theta_g $, $ \mu_g $, $\sigma_g$ and $\beta_g$ leads to  
Alg. \ref{alg:algorithm 2}.
%
%{\setstretch{1.2}

\begin{algorithm}[ht]
%\footnotesize
\small
\caption{MAP-EM for Bayesian mixture model}
\label{alg:algorithm 2}
\begin{algorithmic}[1]
\Require Dataset $\{U^{(n)} = (X^{(n)}, R^{(n)} )  \}_{n=1}^N$, hyperparameters $\Omega$ and number of mixtures $G$.
\State Initialize the parameters $\Theta^{(0)} =  (\theta_1^{(0)} , \dots, \theta_G^{(0)} ) $ and $\Phi^{(0)}  = \{\mu_g^{(0)} , \sigma_g^{(0)} , \beta_g^{(0)} \}_ {g=1}^G$.
\State E-step. For each MTS, evaluate the posteriors $\pi_{g}^{(n)} $ using Eq.~\eqref{eq: p(z|x) posterior} with current parameters $\Phi^{(m)}, \Theta^{(m)}$.
\State M-step. Update parameters using the current posteriors
\begin{align*}
\textstyle
\theta_g^{(m+1)} &= N^{-1} \textstyle \sum_{n=1}^{N} \pi_{g}^{(n)} 
\\
%\sigma_{gv}^2 &= \bigg( N_0 +  \sum_{n=1}^N \sum_{t=1}^T r^{(n)}_v(t) \; \pi_{g}^{(n)}\, \bigg)^{-1} \bigg( N_0 s^2_{v} + \sum_{n=1}^N \sum_{t=1}^T r^{(n)}_v(t) \; \pi^{(n)}_{g} \big(x^{(n)}_v(t) - \mu_{gv}(t)\big)^2 \bigg) 
{\sigma_{gv}^2}^{(m+1)} &= %\big(N_0 +  \textstyle \sum_{n, t=1}^{N, T}  r^{(n)}_v(t) \; \pi_{g}^{(n)}\big)^{-1} \big( N_0 s^2_{v} + \sum_{n, t=1}^{N, T}  r^{(n)}_v(t) \; \pi^{(n)}_{g} \big(x^{(n)}_v(t) - \mu_{gv}(t)\big)^2 \big)
\frac{N_0 s^2_{v} + \sum\limits_{n, t=1}^{N, T}  r^{(n)}_v(t) \; \pi^{(n)}_{g} \big(x^{(n)}_v(t) - \mu_{gv}(t)\big)^2}{N_0 +  \sum_{n, t=1}^{N, T}  r^{(n)}_v(t) \; \pi_{g}^{(n)}}
\\
%\mu_{gv} &= \left( S^{-1}_{v} + \sigma^{-2}_{gv} \textstyle \sum\limits_{n=1}^N  \pi^{(n)}_{g} \text{diag}(r^{(n)}_{v} ) \right)^{-1}  \left( S^{-1}_{v} m_{v} +  \sigma^{-2}_{gv} \textstyle \sum\limits_{n=1}^N  \pi^{(n)}_{g} \text{diag}(r^{(n)}_{v} ) \: x^{(n)}_v \right)   \\
\mu_{gv}^{(m+1)} &= %\big( S^{-1}_{v} + \sigma^{-2}_{gv} \textstyle \sum_{n=1}^N  \pi^{(n)}_{g} \text{diag}(r^{(n)}_{v}) \big)^{-1} \big( S^{-1}_{v} m_{v} +  \sigma^{-2}_{gv} \textstyle \sum_{n=1}^N  \pi^{(n)}_{g} \text{diag}(r^{(n)}_{v} ) \: x^{(n)}_v \big)
\frac{S^{-1}_{v} m_{v} +  \sigma^{-2}_{gv} \textstyle \sum_{n=1}^N  \pi^{(n)}_{g} \text{diag}(r^{(n)}_{v} ) \: x^{(n)}_v}{S^{-1}_{v} + \sigma^{-2}_{gv} \textstyle \sum_{n=1}^N  \pi^{(n)}_{g} \text{diag}(r^{(n)}_{v})}
 \\
  \beta_{gv}^{(m+1)} &=  %\big(c_0 + d_0 -2 + \textstyle \sum_{n=1}^{N} \pi_{g}^{(n)} \big)^{-1} \big(c_0 - 1 + \textstyle \sum_{n=1}^{N} \pi_{g}^{(n)} r^{(n)}_v \big)
  \frac{c_0 - 1 + \textstyle \sum_{n=1}^{N} \pi_{g}^{(n)} r^{(n)}_v}{c_0 + d_0 -2 + \textstyle \sum_{n=1}^{N} \pi_{g}^{(n)}}
\end{align*} 
\State Repeat step 2-3 until convergence.
\Ensure Posteriors $ \Pi^{(n)} \equiv ( \pi_1^{(n)},..,\pi_G^{(n)}) $ and  parameters $\Theta$, $\Phi$.
\end{algorithmic}
\end{algorithm}
%}

\paragraph{The TCK$_{IM}$ kernel}
To compute the TCK$_{IM}$ kernel, we use the mixed mode Bayesian mixture model, described above, as the base model in an ensemble approach. 
Key to ensure that  TCK$_{IM}$ will have statistical advantages (lower variance), computational advantages (less sensitive to local optima) as well as representational advantages (increased expressiveness) compared to the individual base models, is \emph{diversity} and \emph{accuracy}~\cite{Dietterich2000,hansen1990neural}.  In general, this means that the base models should not do the same type of errors and each base model has to perform better than random guessing.  Hence, to ensure diversity, we integrate multiple outcomes of the base model as it is trained under different, randomly chosen, settings (hyperparameters, initialization, subsampling).
In more detail, the number of cluster components for the base models is sampled from  a set of integers $ \mathcal{I}_C = \{I,\dots, I+C\}$. For each number of cluster components $q_2 \in  \mathcal{I}_C $, we apply $Q$ different random initial conditions and sample hyperparameters uniformly as follows: $a_0 \in [0.001, 1]$, $b_0 \in [0.005, 0.2]$, $n_0 \in [0.001, 0.2]$, $c_0, d_0 \in [ 0.1/N, 2/N]$. We let $\mathcal{Q} = \{ q = (q_1,q_2) \: | \: q_1=1,\dots Q, \: q_2 \in \mathcal{I}_C \} $ be the index set keeping track of initial conditions and hyperparameters ($q_1$) as well as the number of components ($q_2$).
Each base model $q$ is trained on a random subset of MTS $\{(X^{(n)}, R^{(n)} )  \}_{n \in \eta(q)}$. To further increase the diversity, for each  $q$, we select random subsets of variables $\mathcal{V}(q)$  as well as random time segments $\mathcal{T}(q)$.
After having trained the individual base models using an embarrasingly parallel procedure, we compute a normalized sum of the inner products of the normalized posterior distributions from each mixture component to build the TCK$_{IM}$ kernel matrix. % Note that, in addition to introducing novel base models to account for informative missingness,  we also modify the kernel  by normalizing the vectors of posteriors to have unit length in the $l_2$-norm. This provides an additional regularization that may increase the generalization capability of the learned model.
Details of the method are presented in Alg.~\ref{alg:algorithm}, whereas Alg.~\ref{alg:algorithm out of sample} describes how to compute the kernel for MTS not seen during training.

% :::::::::::::::::::::: ALGO TCK in-sample ::::::::::::::::::::::
\begin{algorithm}[t!]
%\footnotesize
\small
\caption{TCK$_{IM}$. Training phase.}
\label{alg:algorithm}
\begin{algorithmic}[1]
\Require Training set of MTS $\{(X^{(n)}, R^{(n)} )  \}_{n=1}^N$ , $Q$ initializations, set of integers $\mathcal{I}_C $.
\State Initialize kernel matrix $K = 0_{N \times N}  $.
\For{$q \in \mathcal{Q}$}
\State Compute posteriors $ \Pi^{(n)}(q) \equiv ( \pi_1^{(n)},\dots,\pi_{q_2}^{(n)} )^T $, by fitting a mixed mode mixture model with $q_2$ clusters to the dataset and by randomly selecting:
%{\setstretch{0.4}
\begin{itemize}
\item[i.] hyperparameters $\Omega(q) $%; $a_0 \in [0.001, 1]$, $b_0 \in [0.005, 0.2]$, $n_0 \in [0.001, 0.2]$, $c_0, d_0 \in [ 0.1/N, 2/N]$.
\item[ii.] a time segment $ \mathcal{T}(q)  $ %of length {\small $T_{min} \leq  |\mathcal{T}(q)| \: \leq \: T_{max}$ } 
to extract from each $X^{(n)}$ and $R^{(n)}$,
\item[iv.] a subset of attributes $\mathcal{V}(q)$ %, with cardinality {\small$V_{min} \leq |\mathcal{V}(q)| \leq V_{max}$}, 
to extract from  $X^{(n)}$ and $R^{(n)}$,
\item[vi.] a subset of MTS, $\eta(q) $. %, with {\small$N_{min} \leq |\eta(q)| \leq N$},
\item[vii.] initialization of the mixture parameters $ \Theta(q) $ and $\Phi(q)$.
\end{itemize}
%}
\State Update  $K_{nm} = K_{nm} + \frac{\Pi^{(n)}(q)^T \Pi^{(m)}(q)}{ \| \Pi^{(n)}(q) \| \cdot \| \Pi^{(m)}(q) \| } $.
\EndFor
\Ensure $K$ kernel matrix, time segments $\mathcal{T}(q)  $, subsets of attributes $\mathcal{V}(q)$, subsets of MTS $\eta(q)$, parameters $ \Theta(q)$, $\Phi(q)$  and posteriors $\Pi^{(n)}(q) $.
\end{algorithmic}
\end{algorithm}

% :::::::::::::::::::::::::::::::::::::::::::::::::::::::::::::::::

% :::::::::::::::::::::: ALGO TCK out-sample ::::::::::::::::::::::
\begin{algorithm}[t]
%\footnotesize
\small
\caption{TCK$_{IM}$. Test phase.}
\label{alg:algorithm out of sample}
\begin{algorithmic}[1]
  \Require Test set $\big \{ X^{*(m)} \big \}_{m=1}^M$, time segments $\mathcal{T}(q)  $ subsets $\mathcal{V}(q)$ and $\eta(q)$, parameters $ \Theta(q)$, $\Phi(q)$  and posteriors $\Pi^{(n)}(q) $.
  \State Initialize kernel matrix $K^* = 0_{N \times M} $. 
  \For{$q \in \mathcal{Q}$}
  \State Compute posteriors $\Pi^{*(m)}(q) $, $m=1,\dots,M$ using the mixture parameters $ \Theta(q)$, $\Phi(q)$.
  \State Update  $K^*_{nm} = K^*_{nm} + \frac{\Pi^{(n)}(q)^T \Pi^{*(m)}(q)}{ \| \Pi^{(n)}(q) \| \cdot \| \Pi^{*(m)}(q) \| } $.
  \EndFor
  \Ensure $K^*$ test kernel matrix.
\end{algorithmic}
\end{algorithm}
% :::::::::::::::::::::::::::::::::::::::::::::::::::::::::::::::::

\section{Experiments}
\label{sec: experiments}

To test the performance of the proposed kernel, we considered three clinical datasets of which the characteristics are summarized in Tab.~\ref{tab: char datasets}. The variables and the corresponding missing rates in the three datasets are summarized in Tab.~\ref{tab: blood names}.  A more detailed description of the datasets follows below.

\begin{table}
\centering
{\caption{Description of the three real-world clinical datasets.}\label{tab: char datasets}} 
\begin{tabular}{@{}l|@{\:\:}l@{\:}l@{\:}l@{}}
\hline
\textbf{Dataset} & \textbf{PhysioNet} & \textbf{SSI}& \textbf{AL}\\
\hline
\# of patients & 4000 & 858 & 402 \\
\# attrib (Lab, vital) & 28 (17, 11) & 11 (11, 0) & 11 (7, 4) \\
Length of MTS & 48 (hours) & 10 (days) & 15 (days) \\
Positive class & 874 (21.9\%) & 227 (26.5\%) & 31 (7.7\%) \\
Av. missing rate & 76.6 \%  & 80.7\% &  83\% \\
\hline
\end{tabular}
\end{table}

\begin{table}
\centering
{\caption{List of variables and corresponding missing rates.}\label{tab: blood names}} 
\small
%\begin{tabular}{@{}l@{\:\:}|@{\:\:}l@{\:\:\:}l@{\:\:}l@{\:\:\:}l@{\:\:\:}l@{\:\:\:}l@{}}
\begin{tabular}{l|lll}
\hline
 & \multicolumn{3}{c}{\textbf{Variables}  \textbf{(missing rate) }}\\
\hline
Phys & Albumin (0.99) & ALP (0.98) & AST (0.98)  \\
    & Bilirubin (0.98)   
 & BUN (0.93)  &  Creat. (0.93) \\
 & Diast. BP (0.11)   &  FiO2 (0.84)   & GCS (0.68) \\ 
 & Glucose  (0.93) & HCO3 (0.93) & HCT (0.91) \\
 & HR (0.10) & K (0.93) & Lactate (0.96) \\
 & MAP (0.12) & Mg (0.93) & Na (0.93) \\
 &  PaCO2 (0.88) & PaO2 (0.88)  & Platelets (0.93) \\
 & RespRate (0.76) &  SaO2 (0.96)  & Syst. BP (0.11) \\
 & Temp (0.63) & Urine (0.31) &WBC (0.93) \\
 & pH (0.88) & & \\
 \hline
SSI & Albumin  (0.79)  &  Amylase (0.95) & Creat.  (0.87) \\ 
  & CRP  (0.69) & Glucose  (0.92)  & Hb  (0.65) \\
  &       K (0.71) 
   &  Na  (0.71) & Platelets  (0.92) \\
   & Urea (0.94) & WBC (0.73) &  \\
 \hline
AL &  Albumin (0.80) & Creat. (0.94)  &  CRP (0.77)\\
& Diast. BP (0.76) 
  &  Hb (0.71) &   HR (0.76)  \\
  &  K (0.75)  
   & Na (0.75)  &  Syst. BP (0.76) \\
   & Temp (0.66) &  WBC  (0.80) & \\
\hline
\end{tabular}
\end{table}

\if{
\begin{table*}[tb]
\centering
\caption{List of variables and their corresponding missing rates.} 
\small
\begin{tabular}{@{}l@{\:\:}|@{\:\:}l@{\:\:\:}l@{\:\:}l@{\:\:\:}l@{\:\:\:}l@{\:\:\:}l@{}}
\hline
 & \multicolumn{6}{c}{\textbf{Variables}  \textbf{(missing rate) }}\\
\hline
Phys & Albumin (0.99) & ALP (0.98) & AST (0.98)  & Bilirubin (0.98)   
 & BUN (0.93)  &  Creat. (0.93) \\
 & Diast. BP (0.11)  &  FiO2 (0.84)  
 & GCS (0.68)  & Glucose  (0.93) & HCO3 (0.93) & HCT (0.91) \\
 & HR (0.10) & K (0.93) & Lactate (0.96) & MAP (0.12) 
 & Mg (0.93) & Na (0.93) \\
 &  PaCO2 (0.88) & PaO2 (0.88) 
 & Platelets (0.93) & RespRate (0.76) &  SaO2 (0.96)  & Syst. BP (0.11) \\
 & Temp (0.63) & Urine (0.31) &WBC (0.93) & pH (0.88) & & \\
 \hline
SSI & Albumin  (0.79)  &  Amylase (0.95) & Creat.  (0.87)  
  & CRP  (0.69) & Glucose  (0.92)  & Hb  (0.65) \\
  &       K (0.71) &  Na  (0.71) & Platelets  (0.92)  & Urea (0.94) & WBC (0.73) &   \\
 \hline
AL &  Albumin (0.80) & Creat. (0.94)  &  CRP (0.77) & Diast. BP (0.76) 
  &  Hb (0.71) &   HR (0.76)  \\
   &  K (0.75)  & Na (0.75)  &  Syst. BP (0.76) & Temp (0.66) &  WBC  (0.80) & \\
\hline
\end{tabular}
\label{tab: blood names}
\end{table*}
}\fi

\if{
\begin{table*}[th]
\centering
\caption{Description of the three real-world clinical datasets.} 
\small
\begin{tabular}{l|lll}
\hline
\textbf{Dataset} & \textbf{PhysioNet} & \textbf{Surgical site infection}& \textbf{Anastomosis leakage}\\
\hline
Number of patients & 4000 & 858 & 402 \\
Number of attributes (Lab, vital) & 28 (17, 11) & 11 (11, 0) & 11 (7, 4) \\
Length of MTS & 48 (hours) & 10 (days) & 15 (days) \\
Positive class & 874 (21.9 \%) & 227 (26.5 \%) & 31 (7.7 \%) \\
Average missing rate & 76.6 \%  & 80.7\% &  83\% \\
\hline
\end{tabular}
\label{tab: char datasets}
\end{table*}
}\fi

\paragraph{PhysioNet}  The PhysioNet dataset is collected from the PhysioNet Challenge 2012~\cite{silva2012predicting}. We extracted the first part,  which consists of 4000 patient records of patients from the intensive care units (ICUs) of various hospitals.  Each patient stayed in the ICU for at least 48 hours and the records contain information about both vital signs and blood samples collected over time.  We extracted all measurements taken within 48 hours after admission and aligned the MTS into same-length sequences using an hourly discretization. % If there are two or more measurements per hour, we take the mean.
Variables with a missing rate higher than 99\% were omitted, which led to a total of 28 variables.
The classification task was to predict whether the patient was recovering from surgery, which is a task that also has been considered in other work~\cite{DBLP:journals/corr/ChePCSL16,li2019vs}. 
 
\textbf{Surgical site infection} 
%\paragraph{Case study 2: Detecting infections among patients undergoing colon rectal cancer surgery}
This dataset contains data for 11 blood samples collected postoperatively for patients who underwent major abdominal surgery at the department of gastrointestinal surgery at a Norwegian university hospital in the years 2004-2012. The task considered was to detect surgical site infection (SSI), which is one of the most common types of nosocomial infections~\cite{lewis2013} and  represents up to 30\% of all hospital-acquired infections~\cite{magill2012prevalence}. Patients with no recorded lab tests during the period from postoperative day 1 until day 10 were removed from the cohort, which lead to a final cohort consisting of 858 patients. The average proportion of missing data in the cohort was $80.7 \%$.
To identify the patients in the cohort who developed postoperative SSI and create ground truth labels,
ICD-10 as well as NOMESCO Classification of Surgical Procedures codes related to severe postoperative complications were considered. Patients without these codes who also did not have a mention of the word ``infection'' in any of their postoperative text documents were considered as controls. This lead to a dataset consisting of 227 infected patients and 631 non-infected patients.

%\paragraph{Case study 3: Predicting anastomosis leakage among patients undergoing colon rectal cancer surgery}
\textbf{Anastomosis leakage} 
Anastomosis leakage (AL) is  potentially a serious complication that can occur after colon rectal cancer (CRC) surgery, of which one of the consequences is an increase in 30-day mortality~\cite{snijders2013anastomotic}. 
It is estimated that 5-15\% of the patients who undergo surgery for CRC suffer from AL~\cite{branagan2005prognosis}.  Recent studies have  shown that both unstructured as well structured EHR data such as measurements of blood tests and vital signs could have predictive value for AL~\cite{soguero2016predicting,soguero2014support}. %They followed an imputation approach to deal with missing data and explored whether the relative frequency in the measurements of the structure data (blood tests and vital signs) could have predictive value, assuming that data are MNAR \textbf{CORRECT ???}.
%In this work, we focus just on structured data to leverage the powerful of TCK$_{IM}$ for exploiting the potentially rich information in the missing values, as well as the information from the observed data.
The dataset considered in this work contains only structured data and is collected from the same hospital as the SSI dataset. It contains physiological data (blood tests and vital signs) for the period from the day after surgery to day 15 for 402 patients who underwent CRC surgery. A total of 31 of these patients got AL after surgery, but there is no information available about exactly when it  happened. The classification task considered here was to detect which of the patients got AL.
%Our goal is to predict AL based on a multivariate time series composed by nine different blood tests, namely, albumin, C-Reactive Protein (CRP), glucose, hemoglobin, potassium, creatinine, leukocytes, sodium, and thrombocytes; and by four vital signs, namely, temperature, systolic and diastolic blood pressure (B and pulse. 

\paragraph{Experimental setup}
We considered the following experimental setup.
We performed kernel principal component analysis (KPCA)~\cite{scholkopf1997kernel} using the proposed TCK$_{IM}$ and then trained a kNN-classifier in the low dimensional space.  The dimensionality of the KPCA-representation was set to 3 to also be able to visualize the embeddings, whereas we used 5-fold cross validation to set the number of neighbors $k$ for the kNN-classifier. 
Performance was measured in terms of F1-score, sensitivity and specificity.
Sensitivity is the fraction of correctly classified cases, whereas specificity is the fraction of controls that are correctly
classified as negative. F1-score is the harmonic mean of precision and sensitivity, where precision is the fraction of actual positives among all those that are classified as positive
cases.

We compared the performance of the proposed kernel to four baseline kernels, namely the linear kernel (Lin), the global alignment kernel (GAK)~\cite{cuturi2011fast}, LPS and TCK.
GAK is a positive semi-definite kernel formulation of the widely used, but non-metric, time series similarity measure called \emph{dynamic time warping} (DTW) \cite{Berndt:1994:UDT:3000850.3000887}.  It has two hyperparameters, namely  the kernel bandwidth and the triangular parameter, which have to be set by the user and it does not naturally deal with missing data and incomplete datasets, and therefore also requires a preprocessing step involving imputation. Therefore, we created a complete dataset using mean imputation for Lin and GAK (initial experiments showed that mean imputation worked better than last observation carried forward). In accordance with~\cite{Cuturi}, for GAK we set the bandwidth $\sigma$ to 0.1 times the median  distance of all MTS  in the training set scaled by the square root of the median length of all MTS, and the triangular parameter to 0.2 times the median length (Frobenius norm) of all MTS. 
In order to design baseline kernels that can exploit informative missingness, we also created baselines (referred to as Lin$_{IM}$, GAK$_{IM}$ and LPS$_{IM}$) by concatenating the binary indicator MTS $R^{(n)}$ to $X^{(n)}$.
LPS was run with default hyperparameters using the implementation provided by~\cite{Baydogan}, with the exception that the minimal segment length was adjusted to account for the relatively short MTS in the datasets. 
For the TCK$_{IM}$ we let $Q = 15$ and $\mathcal{I}_C = \{N/200,\dots, N/200+20\}$, and, likewise, TCK was run with $Q = 15$ and $C = 20$.  For all methods, except LPS which do not require standardization, we standardized each attribute to zero mean and unit standard deviation.

For PhysioNet and SSI, we did 5-fold cross validation to measure performance. The AL dataset is, however, highly imbalanced and therefore we employed an undersampling strategy by randomly sampling two patients from the negative class per positive case. 20\% of this dataset was then set aside as a test set. This entire process was repeated 10 times and we reported mean and standard errors of the performance measures. The AL dataset is small, and for that reason the hyperparameters of the methods had to be adjusted accordingly. For TCK$_{IM}$ we let $Q = 10$ and $\mathcal{I}_C = \{2,3\}$.

\if{
\begin{table*}[t]
    \centering
        \caption{Performance (mean $\pm$ se) for different kernels and datasets.}
       % \small
    \begin{tabular}{@{}l|llcccc@{}}
    \hline
\textbf{Dataset}  & & \textbf{Kernel} & \textbf{Sensitivity} & \textbf{Specificity} & \textbf{F1-score} & \textbf{Accuracy} \\
     \hline
\multirow{8}{*}{Phys.} 
& Impute &   Lin  & 0.329 $\pm$ 0.049  & 0.812 $\pm$ 0.007  &  0.328 $\pm$ 0.052  &  0.707  $\pm$ 0.007   \\
& &   GAK  & 0.313 $\pm$ 0.021 & 0.801 $\pm$ 0.019  & 0.309 $\pm$ 0.012 & 0.694  $\pm$ 0.013   \\
 & Ignore &   LPS  & 0.511 $\pm$ 0.069   & 0.948 $\pm$ 0.010 & 0.600 $\pm$ 0.068 & 0.852 $\pm$ 0.016 \\
&  &   TCK  & 0.411 $\pm$ 0.053  & 0.833 $\pm$ 0.022 & 0.408 $\pm$ 0.049 & 0.740 $\pm$ 0.017   \\
 & Informative &   Lin$_{IM}$  & 0.556 $\pm$ 0.054 & 0.939 $\pm$ 0.013  & 0.625 $\pm$ 0.046  & 0.855  $\pm$ 0.013    \\
&  &   GAK$_{IM}$   & 0.566 $\pm$ 0.032 & 0.941  $\pm$ 0.010 & 0.636  $\pm$ 0.038  & 0.858 $\pm$ 0.007   \\ 
 & &   LPS$_{IM}$  & 0.611  $\pm$ 0.060  & 0.939 $\pm$ 0.009  & 0.667 $\pm$ 0.041  & 0.867  $\pm$ 0.016    \\
&  &   TCK$_{IM}$ & \textbf{0.699   $\pm$ 0.034}   & \textbf{0.980  $\pm$ 0.007}  & \textbf{0.789   $\pm$  0.033} &\textbf{0.915   $\pm$ 0.011}  \\ 
   \hline
\multirow{8}{*}{SSI}  
& Impute & Lin  & 0.480 $\pm$ 0.041 & 0.878 $\pm$ 0.041 & 0.529 $\pm$ 0.072 & 0.772 $\pm$ 0.040  \\  
& &   GAK & 0.639 $\pm$ 0.064 & 0.921 $\pm$ 0.036 & 0.687 $\pm$ 0.025  & 0.846 $\pm$ 0.015 \\
& Ignore & LPS   & 0.687 $\pm$ 0.044 & 0.929 $\pm$ 0.040 & 0.730 $\pm$ 0.028& 0.865 $\pm$ 0.020  \\
& &   TCK  & 0.683 $\pm$ 0.071 & 0.922 $\pm$ 0.025 & 0.719 $\pm$ 0.069  & 0.859  $\pm$ 0.034   \\

& Informative &  Lin$_{IM}$  & 0.700 $\pm$ 0.012 & \textbf{0.944 $\pm$ 0.030} & 0.755 $\pm$ 0.024 & 0.879 $\pm$ 0.012  \\
& &   GAK$_{IM}$ & 0.718 $\pm$ 0.073 & 0.940 $\pm$ 0.026  & 0.761 $\pm$ 0.036 & 0.881 $\pm$ 0.015 \\
& &   LPS$_{IM}$   & 0.652 $\pm$ 0.056 & 0.925 $\pm$ 0.015 & 0.701 $\pm$ 0.036& 0.853 $\pm$ 0.015  \\
& &   TCK$_{IM}$   & \textbf{0.775 $\pm$ 0.017} & 0.929 $\pm$ 0.010 & \textbf{0.786 $\pm$ 0.015} & \textbf{0.888 $\pm$ 0.009} \\
 \hline
\multirow{8}{*}{AL}  
& Impute & Lin  &  0.414 $\pm$ 0.273  & 0.907 $\pm$ 0.119 & 0.468 $\pm$ 0.240  & 0.742  $\pm$ 0.060    \\
%std 
& &   GAK  & 0.428 $\pm$ 0.223 & \textbf{0.935 $\pm$ 0.078}  & 0.520 $\pm$ 0.206  & 0.766 $\pm$ 0.065   \\
& Ignore &  LPS  &  \textbf{0.843 $\pm$  0.105} & 0.835  $\pm$  0.075 & 0.776 $\pm$ 0.058  & 0.838 $\pm$ 0.046   \\
&  &   TCK  &  0.700 $\pm$ 0.147 & 0.886 $\pm$ 0.107 &  0.722 $\pm$ 0.080 & 0.823 $\pm$ 0.050   \\
& Informative &   Lin$_{IM}$  & 0.742 $\pm$ 0.221  & 0.864 $\pm$ 0.097  & 0.724  $\pm$ 0.097  & 0.823  $\pm$ 0.039    \\

& &   GAK$_{IM}$  & 0.728 $\pm$ 0.227  & 0.928  $\pm$ 0.047  & 0.759 $\pm$ 0.156  & \textbf{0.861 $\pm$ 0.065}  \\ 
& &   LPS$_{IM}$  &  0.800 $\pm$ 0.120 & 0.864 $\pm$ 0.091  & 0.773 $\pm$ 0.091  & 0.843  $\pm$ 0.067   \\ 
& &   TCK$_{IM}$   & \textbf{0.843 $\pm$ 0.184} & 0.864 $\pm$ 0.095 & \textbf{0.792 $\pm$ 0.115} & 0.857 $\pm$ 0.072 \\
    \hline
    \end{tabular}
    \label{tab:Results}
\end{table*}
}\fi

\section{Results and discussion}
\label{Sec:Results}

\begin{table}[t]
    \centering
        \caption{Performance (mean $\pm$ se) on 3 datasets.}
        \small
    \begin{tabular}{@{}l@{\:}|l@{\:\:}c@{\:\:\:\:}c@{\:\:\:\:}c@{}}
    \hline
 &  \textbf{Kernel} & \textbf{Sensitivity} & \textbf{Specificity} & \textbf{F1-score}  \\
     \hline
\multirow{8}{*}{Phys.} 
&  Lin  & 0.329 $\pm$ 0.049  & 0.812 $\pm$ 0.007  &  0.328 $\pm$ 0.052   \\
&    GAK  & 0.313 $\pm$ 0.021 & 0.801 $\pm$ 0.019  & 0.309 $\pm$ 0.012   \\
 &   LPS  & 0.511 $\pm$ 0.069   & 0.948 $\pm$ 0.010 & 0.600 $\pm$ 0.068  \\
&     TCK  & 0.411 $\pm$ 0.053  & 0.833 $\pm$ 0.022 & 0.408 $\pm$ 0.049    \\
 &    Lin$_{IM}$  & 0.556 $\pm$ 0.054 & 0.939 $\pm$ 0.013  & 0.625 $\pm$ 0.046     \\
&     GAK$_{IM}$   & 0.566 $\pm$ 0.032 & 0.941  $\pm$ 0.010 & 0.636  $\pm$ 0.038     \\ 
 &    LPS$_{IM}$  & 0.611  $\pm$ 0.060  & 0.939 $\pm$ 0.009  & 0.667 $\pm$ 0.041     \\
&    TCK$_{IM}$ & \textbf{0.699   $\pm$ 0.034}   & \textbf{0.980  $\pm$ 0.007}  & \textbf{0.789   $\pm$  0.033} \\ 
   \hline
\multirow{8}{*}{SSI}  
& Lin  & 0.480 $\pm$ 0.041 & 0.878 $\pm$ 0.041 & 0.529 $\pm$ 0.072  \\  
&   GAK & 0.639 $\pm$ 0.064 & 0.921 $\pm$ 0.036 & 0.687 $\pm$ 0.025   \\
& LPS   & 0.687 $\pm$ 0.044 & 0.929 $\pm$ 0.040 & 0.730 $\pm$ 0.028 \\
&    TCK  & 0.683 $\pm$ 0.071 & 0.922 $\pm$ 0.025 & 0.719 $\pm$ 0.069    \\
&   Lin$_{IM}$  & 0.700 $\pm$ 0.012 & \textbf{0.944 $\pm$ 0.030} & 0.755 $\pm$ 0.024  \\
&    GAK$_{IM}$ & 0.718 $\pm$ 0.073 & 0.940 $\pm$ 0.026  & 0.761 $\pm$ 0.036  \\
&    LPS$_{IM}$   & 0.652 $\pm$ 0.056 & 0.925 $\pm$ 0.015 & 0.701 $\pm$ 0.036 \\
&   TCK$_{IM}$   & \textbf{0.775 $\pm$ 0.017} & 0.929 $\pm$ 0.010 & \textbf{0.786 $\pm$ 0.015} \\
 \hline
\multirow{8}{*}{AL}  
& Lin  &  0.414 $\pm$ 0.273  & 0.907 $\pm$ 0.119 & 0.468 $\pm$ 0.240     \\
%std 
&   GAK  & 0.428 $\pm$ 0.223 & \textbf{0.935 $\pm$ 0.078}  & 0.520 $\pm$ 0.206   \\
&   LPS  &  \textbf{0.843 $\pm$  0.105} & 0.835  $\pm$  0.075 & 0.776 $\pm$ 0.058    \\
&     TCK  &  0.700 $\pm$ 0.147 & 0.886 $\pm$ 0.107 &  0.722 $\pm$ 0.080   \\
&   Lin$_{IM}$  & 0.742 $\pm$ 0.221  & 0.864 $\pm$ 0.097  & 0.724  $\pm$ 0.097    \\
&   GAK$_{IM}$  & 0.728 $\pm$ 0.227  & 0.928  $\pm$ 0.047  & 0.759 $\pm$ 0.156   \\ 
&    LPS$_{IM}$  &  0.800 $\pm$ 0.120 & 0.864 $\pm$ 0.091  & 0.773 $\pm$ 0.091   \\ 
&   TCK$_{IM}$   & \textbf{0.843 $\pm$ 0.184} & 0.864 $\pm$ 0.095 & \textbf{0.792 $\pm$ 0.115}  \\
    \hline
    \end{tabular}
    \label{tab:Results}
\end{table}

Tab.~\ref{tab:Results} shows the performance of the TCK$_{IM}$  kernel, as well as the baseline methods, on the three real-world datasets. %We performed 5-fold cross validation and reported results in terms of sensitivity, specificity, F1-score and accuracy. 
The first thing to notice is that the two kernels (Lin and GAK) that rely on imputation consistently perform much worse than the other kernels in terms of both F1-score across all datasets. We also note that these methods achieve a relatively high specificity. However, this is because they put too many patients in the negative class, which also leads to a high false negative rate and, consequently, a low sensitivity. 
The reasons could be that the imputation  methods introduce biases and that the missingness mechanism is ignored. %\st{We also note that the standard errors for these methods are very high.}

\begin{figure*}[t]
\centering
\includegraphics[width=0.95\linewidth]{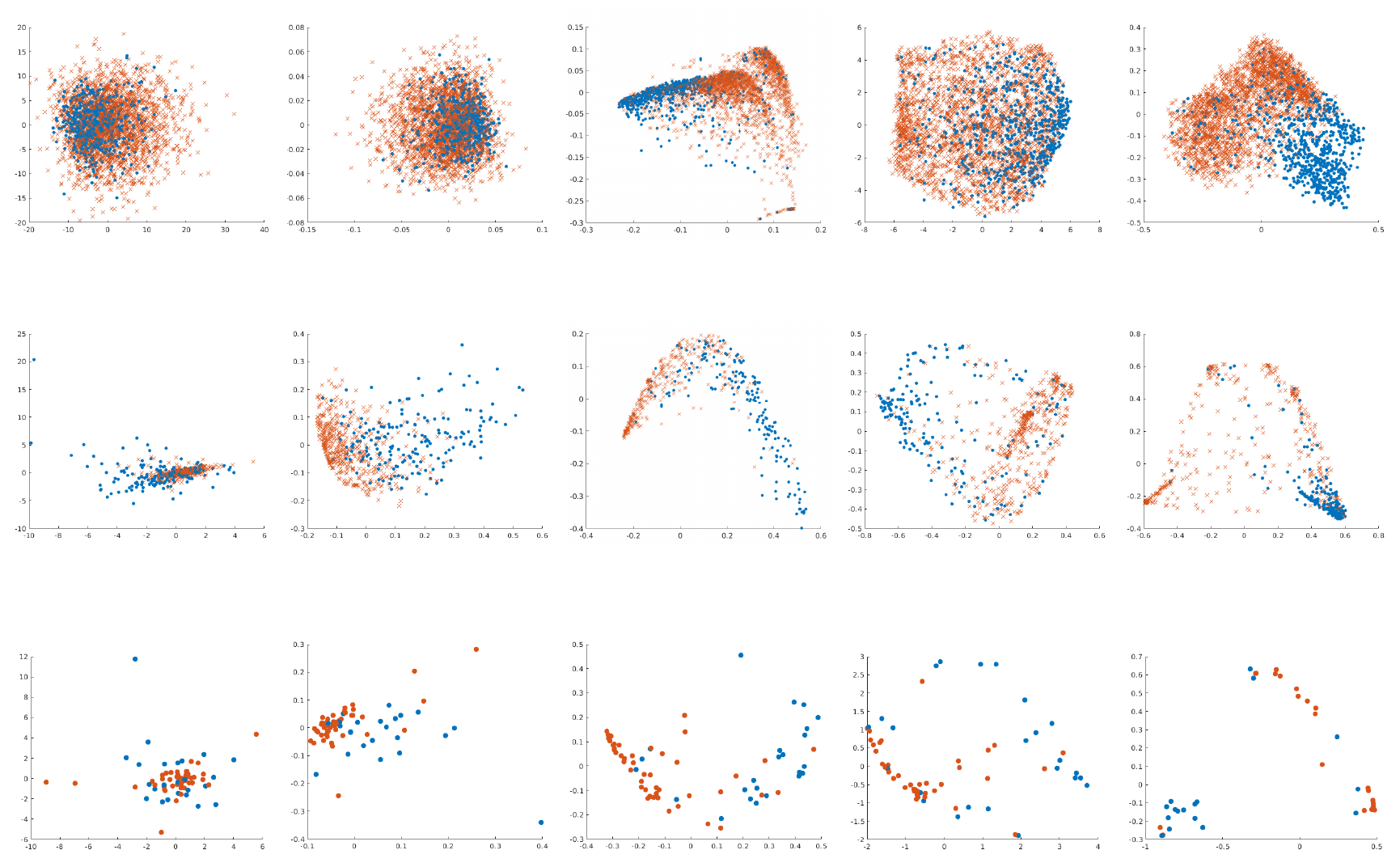} % Reduce the figure size so that it is slightly narrower than the column. Don't use precise values for figure width.This setup will avoid overfull boxes. 
    \caption{KPCA plots obtained using Lin, GAK, LPS, TCK and TCK$_{IM}$, respectively. Row 1: Physio., row 2: SSI, row 3: AL. %The blue dots represent the positive class.
    }
    \label{fig: KPCA plots}
\end{figure*}

The two kernels, TCK and LPS, that naturally handle the missing data perform better than the kernels that rely on imputation for all three datasets.
TCK and LPS perform quite similarly across all 3 evaluation metrics for the SSI dataset, whereas LPS outperforms TCK on the PhysioNet and AL dataset.  
These methods probably perform better than the imputation methods because ignoring the missingness introduces less bias than replacing missing values with biased estimates.
The performance of the baselines that account for informative missingness, Lin$_{IM}$ and GAK$_{IM}$, is considerably better than Lin and GAK, respectively, for all datasets.  LPS$_{IM}$ also performs better than LPS on the PhysioNet datasets, whereas the performance of these two baselines is more or less equal on the two other datasets.
The proposed TCK$_{IM}$ performs considerably better than all baselines, and in particular compared to TCK (the kernel which it is an improvement of) for the PhysioNet and SSI datasets in terms of F1-score, and it performs better or comparable than the other kernels on the AL dataset.
This demonstrates that the missing patterns in clinical time series are often informative and the TCK$_{IM}$ can exploit this information very efficiently.

Fig.~\ref{fig: KPCA plots} shows the KPCA embeddings obtained
using five kernels (Lin, GAK, LPS, TCK and TCK$_{IM}$). In general, the
classes are more separated in the representations obtained using TCK$_{IM}$ than in the other representations.

\if{
\begin{figure*}[!t]
    \centering
        \subfigure
    {
        \includegraphics[trim = {42mm 85mm 42mm 57mm}, clip, width=0.187\linewidth]{figures/physio_lin.pdf}
    }%
      \subfigure
    {
        \includegraphics[trim = {42mm 85mm 42mm 57mm}, clip, width=0.187\linewidth]{figures/physio_gak.pdf}
    }%
        \subfigure
    {
        \includegraphics[trim = {42mm 85mm 42mm 57mm}, clip, width=0.187\linewidth]{figures/physio_lps.pdf}
    }%
        \subfigure
    {
        \includegraphics[trim = {42mm 85mm 42mm 57mm}, clip, width=0.187\linewidth]{figures/physio_tck.pdf}
    }%
    \subfigure
    {
        \includegraphics[trim = {42mm 85mm 42mm 57mm}, clip, width=0.187\linewidth]{figures/physio_tckim.pdf}
    }  
            \subfigure
    {
        \includegraphics[trim = {42mm 85mm 42mm 57mm}, clip, width=0.187\linewidth]{figures/SSI_lin.pdf}
    }%
      \subfigure
    {
        \includegraphics[trim = {42mm 85mm 42mm 57mm}, clip, width=0.187\linewidth]{figures/SSI_gak.pdf}
    }%
        \subfigure
    {
        \includegraphics[trim = {42mm 85mm 42mm 57mm}, clip, width=0.187\linewidth]{figures/SSI_lps.pdf}
    }%
        \subfigure
    {
        \includegraphics[trim = {42mm 85mm 42mm 57mm}, clip, width=0.187\linewidth]{figures/SSI_tck.pdf}
    }%
    \subfigure
    {
        \includegraphics[trim = {42mm 85mm 42mm 57mm}, clip, width=0.187\linewidth]{figures/SSI_tckim.pdf}
    }  
            \subfigure
    {
        \includegraphics[trim = {42mm 85mm 42mm 57mm}, clip, width=0.187\linewidth]{figures/AL_lin.pdf}
    }%
      \subfigure
    {
        \includegraphics[trim = {42mm 82mm 42mm 55mm}, clip, width=0.187\linewidth]{figures/AL_gak.pdf}
    }%
        \subfigure
    {
        \includegraphics[trim = {42mm 82mm 42mm 55mm}, clip, width=0.187\linewidth]{figures/AL_lps.pdf}
    }%
        \subfigure
    {
        \includegraphics[trim = {42mm 85mm 42mm 57mm}, clip, width=0.187\linewidth]{figures/AL_tck.pdf}
    }%
    \subfigure
    {
        \includegraphics[trim = {42mm 85mm 42mm 57mm}, clip, width=0.187\linewidth]{figures/AL_tckim.pdf}
    }  
    \caption{Plot of KPCA representations obtained using Lin, GAK, LPS, TCK and TCK$_{IM}$, respectively. Row 1: PhysioNet, row 2: SSI, row 3: AL. The blue dots represent the positive class.}
    \label{fig: KPCA plots}
\end{figure*}

\begin{figure*}[!t]
    \centering
        \subfigure[Linear]
    {
        \includegraphics[trim = {42mm 85mm 42mm 57mm}, clip, width=0.187\linewidth]{figures/physio_lin.pdf}
    }%
      \subfigure[GAK]
    {
        \includegraphics[trim = {42mm 85mm 42mm 57mm}, clip, width=0.187\linewidth]{figures/physio_gak.pdf}
    }%
        \subfigure[LPS]
    {
        \includegraphics[trim = {42mm 85mm 42mm 57mm}, clip, width=0.187\linewidth]{figures/physio_lps.pdf}
    }%
        \subfigure[TCK]
    {
        \includegraphics[trim = {42mm 85mm 42mm 57mm}, clip, width=0.187\linewidth]{figures/physio_tck.pdf}
    }%
    \subfigure[TCK$_{IM}$]
    {
        \includegraphics[trim = {42mm 85mm 42mm 57mm}, clip, width=0.187\linewidth]{figures/physio_tckim.pdf}
    }  
            \subfigure[Linear]
    {
        \includegraphics[trim = {42mm 85mm 42mm 57mm}, clip, width=0.187\linewidth]{figures/SSI_lin.pdf}
    }%
      \subfigure[GAK]
    {
        \includegraphics[trim = {42mm 85mm 42mm 57mm}, clip, width=0.187\linewidth]{figures/SSI_gak.pdf}
    }%
        \subfigure[LPS]
    {
        \includegraphics[trim = {42mm 85mm 42mm 57mm}, clip, width=0.187\linewidth]{figures/SSI_lps.pdf}
    }%
        \subfigure[TCK]
    {
        \includegraphics[trim = {42mm 85mm 42mm 57mm}, clip, width=0.187\linewidth]{figures/SSI_tck.pdf}
    }%
    \subfigure[TCK$_{IM}$]
    {
        \includegraphics[trim = {42mm 85mm 42mm 57mm}, clip, width=0.187\linewidth]{figures/SSI_tckim.pdf}
    }  
            \subfigure[Linear]
    {
        \includegraphics[trim = {42mm 85mm 42mm 57mm}, clip, width=0.187\linewidth]{figures/AL_lin.pdf}
    }%
      \subfigure[GAK]
    {
        \includegraphics[trim = {42mm 85mm 42mm 57mm}, clip, width=0.187\linewidth]{figures/AL_gak.pdf}
    }%
        \subfigure[LPS]
    {
        \includegraphics[trim = {42mm 85mm 42mm 57mm}, clip, width=0.187\linewidth]{figures/AL_lps.pdf}
    }%
        \subfigure[TCK]
    {
        \includegraphics[trim = {42mm 85mm 42mm 57mm}, clip, width=0.187\linewidth]{figures/AL_tck.pdf}
    }%
    \subfigure[TCK$_{IM}$]
    {
        \includegraphics[trim = {42mm 85mm 42mm 57mm}, clip, width=0.187\linewidth]{figures/AL_tckim.pdf}
    }  
    \caption{Plot of the KPCA representations obtained using five different kernels. (a)-(e): PhysioNet, (f)-(j): surgical site infection, (k)-(o): anastomosis leakage.}
    \label{fig: KPCA plots}
\end{figure*}
}\fi

\paragraph{Limitations, future work and conclusions}
%\label{Sec:Conclusion}
%\textbf{update}: lack of labels.
%\textbf{mention what happens if missingness is not informative.}
In this paper, we presented a MTS kernel capable of exploiting
informative missingness along with the temporal dependencies in the observed data. We showed that TCK$_{IM}$ can learn good representations that can be exploited both in supervised and unsupervised tasks, even when the percentage of missing data is high. In this work, the representations learned using TCK$_{IM}$ were evaluated visually (Fig.~\ref{fig: KPCA plots}) and using a supervised scheme by training a classifier (Tab.~\ref{tab:Results}). The experimental results suggested that TCK$_{IM}$ achieved superior performances compared to baselines on three real datasets.  

The experiments presented in this work focused on binary classification tasks, both of patients at the ICU and patients who had undergone colonrectal cancer surgery.
However, we believe that TCK$_{IM}$ is also a very good choice in applications where there is a lack of labels, which often is the case in medical applications, thanks to the ensemble learning strategy that makes the kernel robust to hyperparameter choices. In fact, since it is a kernel, it can be used in many different applications, including classification as well as clustering tasks, benefiting from the vast body of work in the field of kernel methods. In future work, we would like to test TCK$_{IM}$ in a realistic unsupervised task from the medical domain. %However, lack of labels and massive missing data might also occur in other application areas and we therefore foresee that the proposed methods could have other uses.

A limitation of TCK$_{IM}$ is that it is only designed for MTS of the same length. In further work, we would therefore like to design a time series cluster kernel that can also deal with varying length MTS.
It should also be pointed out that if the missing patterns are not informative, i.e. the missingness is not correlated the particular medical condition(s) of interest, the performance gain of TCK$_{IM}$ compared to TCK is low. It is therefore an advantage if the user has some understanding about the underlying missingness mechanism in the data. On the other hand, our experiments on benchmark datasets (see Appendix) demonstrate that in cases when the missingness mechanism is almost ignorable (low correlation between missing patterns and labels), the performance of TCK$_{IM}$ is not worse than TCK.

\section{Acknowledgement}
The authors would like to thank K. Hindberg  for assistance on extraction of EHR data and the physicians A. Revhaug, R.-O. Lindsetmo and K. M. Augestad for helpful guidance throughout the study.

\section{Appendix -- Synthetic benchmark datasets}

To test how well TCK$_{IM}$ performs for  a  varying degree of informative missingness,  we  generated in total 16 synthetic datasets by randomly injecting missing data into 4 MTS benchmark datasets. The characteristics of the datasets are described in Tab.~\ref{tab: benchmark description}. 
We transformed all MTS in each dataset to the same length, $T$, where T is given by
$
    T = \left \lceil T_{max} / \left \lceil T_{max}/25 \right \rceil \right \rceil.
$
Here, $ \lceil \: \rceil$ is the ceiling operator and $T_{max}$ is the length of the longest MTS in the original dataset.

\begin{table}[h]
\centering
\caption{Characteristics of the benchmark datasets. Attr is the number of attributes, Train and Test the number of training and test samples. $N_c$ is the number of classes, $T_{min}$ and $T_{max}$ the length of shortest and longest MTS in the dataset,  whereas $T$ is the length of the MTS after the transformation.}\label{tab: benchmark description}
\small
%\footnotesize
\begin{tabular}{@{}l@{\:}c@{\:}c@{\:}c@{\:\:}c@{\:}c@{}c@{\:}c@{\:}c@{}}
\hline
\textbf{Datasets} & \textbf{Attr} &  \textbf{Train}  & \textbf{Test}  &  $N_c$  & $T_{min}$ & $T_{max}$ & $T$ & \textbf{Source }\\
\hline
uWave & 3 & 200 & 4278 & 8 & 315 & 315 & 25 & \cite{bagnall2018uea} \\
%uWave$_{25}$ & 3 & 200 & 4278 & 8 & 315 & 315 & 25 & UCR \\
Char.tra. & 3 & 300 & 2558 & 20 & 109 & 205 & 23 & \cite{bagnall2018uea}  \\ %UCI~\cite{Lichman:2013} \\
Wafer & 6 & 298 & 896 & 2 & 104 & 198 & 25 & ~\cite{Olszewski} \\
Jap.vow. & 12 & 270 & 370 & 9 & 7 & 29 & 15 & \cite{bagnall2018uea} \\ % \\
\hline
\end{tabular}
\end{table}

\paragraph{Datasets}
The following procedure was used to create 8 synthetic datasets with missing data from the Wafer and Japanese vowels datasets. We randomly sampled a number $c_v \in \{-1, 1\} $ for each attribute $v \in \{1,\dots, V\} $,  where $c_v =1$ indicates that the attribute and the labels are positively correlated and $c_v =-1$ negatively correlated. 
Thereafter, we  sampled a missing rate $\gamma_{nv}$ from $\mathcal{U}[  0.3 + E \cdot c_v \cdot  (y^{(n)}-1), 0.7 + E \cdot c_v \cdot (y^{(n)}-1)]$ for each MTS $X^{(n)}$ and attribute.  The parameter $E$ was tuned such that the Pearson correlation (absolute value) between the missing rates for the attributes $\gamma_v$  and the labels $y^{(n)}$ took the values  $\{ 0.2, \: 0.4, \: 0.6, \: 0.8\}$, respectively. By doing so, we could control the amount of informative missingness and because of the way we sampled $\gamma_{nv}$, the missing rate in each dataset was around 50\% independently of the Pearson correlation. 

Further, the following procedure was used to create 8 synthetic datasets from the uWave  and Character trajectories datasets, which both consist of only 3 attributes. We randomly sampled a number $c_v \in \{-1, 1\} $ for each attribute $v \in \{1,\dots, V\} $.  Attribute(s) with $c_v = -1$ became negatively correlated with the labels by sampling  $\gamma_{nv}$  from  
$\mathcal{U}[ 0.7 - E \cdot  (y^{(n)}-1), 1 - E  \cdot (y^{(n)}-1)]$, whereas the attribute(s) with $c_v = 1$ became positively correlated with the labels by sampling  $\gamma_{nv}$  from  
$\mathcal{U}[ 0.3 + E \cdot  (y^{(n)}-1), 0.6 + E  \cdot (y^{(n)}-1)]$. The parameter $E$ was computed in the same way as above.
Then, we computed the mean of each attribute $\mu_v$ over the complete dataset and let each element with $  x^{(n)}_v(t) > \mu_v$ be missing with probability $\gamma_{nv}$. This means  that the probability of being missing is dependent on the value of the missing element, i.e. the missingness mechanism is MNAR within each class. Hence, this type of informative missingness is not the same as the one we created for the Wafer and Japanese vowels datasets. 

\paragraph{Baselines}
Three baseline models were created. 
The first baseline, namely ordinary TCK, ignores the missingness mechanism. 
We created a second baseline, refered to as TCK$_B$, in which the missing patterns we modeled naively by concatenating the binary missing indicator MTS $R$ to the MTS $X$ and creating a new MTS $U$ with $2V$ attributes. Then, ordinary TCK was trained on the datasets consisting of $\{U^{(n)}\}$. 
In the third baseline, TCK$_0$, we investigated how well informative missingness can be captured by imputing zeros for the missing values and then training the TCK on the imputed data.

\begin{table}[!th]
\centering
\caption{Performance (accuracy) on the 16 synthetic datasets}
\label{tab: im synthetic results}
\small
\begin{tabular}{|l|l|llll|}
\hline
Dataset & Corr. &  TCK  & TCK$_{B}$ & TCK$_{0}$ &  TCK$_{IM}$   \\
\hline
% &  \multicolumn{4}{c|}{\textbf{Wafer}} & \multicolumn{4}{c}{\textbf{Japanese vowels}} \\
%\hline
%0 & 0.958  & 0.954  &\textbf{ 0.959}  & 0.943  &  \textbf{0.938} &  0.930 & \textbf{0.938} &  0.935   \\
\multirow{4}{*}{Wafer} & 0.2  & 0.951  & 0.951  & 0.951  & \textbf{0.955}  \\
& 0.4 & \textbf{0.961}  & 0.953  & 0.955  & \textbf{0.961}    \\
& 0.6 & 0.961  & 0.900  & 0.965  & \textbf{0.996}     \\
& 0.8  & 0.958  & 0.893  & 0.963  & \textbf{1.000}    \\
% 0.967567567567568	0.924324324324324	0.935135135135135	0.921621621621622
%&  0.916 &   0.973   &   0.975  &  \textbf{0.978}  \\
\hline
\multirow{4}{*}{Japan. vow.} & 0.2 & 0.938 &  \textbf{0.954} & 0.951 &  0.940   \\
 & 0.4 &  0.932 &  0.938 & 0.938 &  \textbf{0.941}  \\
 & 0.6 &  0.922 &  0.946 & 0.924 &  \textbf{0.962} \\
 &0.8 &  0.922 &   0.924  &   0.935  &  \textbf{0.968} \\
\hline
%&  \multicolumn{4}{c|}{\textbf{uWave}} & \multicolumn{4}{c}{\textbf{Character trajectories}} \\
\multirow{4}{*}{uWave} & 0.2 & 0.763 & 0.457 &  0.755 & \textbf{0.841} \\
& 0.4 & 0.807 & 0.587 & 0.813 & \textbf{0.857} \\
& 0.6 & 0.831 & 0.674 & 0.837 & \textbf{0.865} \\
& 0.8 & 0.834 & 0.699 & 0.844 & \textbf{0.884} \\
\hline
\multirow{4}{*}{Char. Traj.} & 0.2 & \textbf{0.854} & 0.742 & 0.847 & 0.851 \\
 & 0.4 & 0.851 & 0.788 & 0.842 & \textbf{0.867}\\
& 0.6  & 0.825 & 0.790 & 0.824 & \textbf{0.871} \\
& 0.8 & 0.839 & 0.707 & 0.853 & \textbf{0.901} \\
\hline
\end{tabular}
\end{table}

\paragraph{Results}
Tab.~\ref{tab: im synthetic results} shows the performance of the proposed TCK$_{IM}$ and the three baselines for all of the 16 synthetic datasets.
We see that the proposed TCK$_{IM}$ achieves the best accuracy for 14 out of 16 datasets and is the only method which consistently has the expected behaviour, namely that the accuracy increases as the correlation between missing values and class labels increases. It can also be seen that the performance of TCK$_{IM}$ is similar to TCK when the amount of information in the missing patterns is low, whereas TCK is clearly outperformed when the informative missingness is high. This demonstrates that TCK$_{IM}$ can effectively exploit informative missingness.

\bibliographystyle{aaai}
\fontsize{9.0pt}{10.0pt} \selectfont
%{\small
\bibliography{biblio}
%}
%}

\end{document}